\documentclass[letterpaper, 10 pt, conference]{ieeeconf}  

\IEEEoverridecommandlockouts                              

\usepackage{color}
\usepackage{graphicx}
\usepackage{xcolor}
\usepackage{amsmath,amssymb,amsfonts,mathrsfs}
\usepackage[colorlinks=true, allcolors=blue]{hyperref}
\usepackage[caption=false,font=footnotesize]{subfig}
\usepackage{stfloats}
\usepackage{algorithm}
\usepackage{algorithmic}
\usepackage{multirow}
\usepackage{lipsum}
\usepackage{pgf}
\usepackage{tikz}
\usetikzlibrary{arrows.meta}
\usepackage{tcolorbox}
\usepackage{xspace}
\usepackage{siunitx}
\usepackage{booktabs}

\usetikzlibrary{arrows,automata,calc,petri,shapes,positioning}
\usetikzlibrary{shapes.multipart}
\usetikzlibrary{shapes.misc}
\tikzset{automatonFig/.style={->,>=stealth',shorten >=1pt,auto,node distance=1.7cm,
                    thick,initial text={}}}
\tikzset{forbidding cross/.style={cross out,draw,-,
         minimum size=0.8cm,thin}}  
         
\definecolor{aliceblue}{rgb}{0.94, 0.97, 1.0}
\definecolor{lightyellow}{rgb}{1.0, 1.0, 0.88}
\definecolor{definitioncolor}{rgb}{0.94, 0.97, 1.0}
\definecolor{examplecolor}{rgb}{1.0, 1.0, 0.88}
\definecolor{publicationboxcolor}{rgb}{0.94, 1.0, 0.94}
\definecolor{rqcolor}{rgb}{1.0, 0.89, 0.88}
\definecolor{anti-flashwhite}{rgb}{0.95, 0.95, 0.96}
\definecolor{antiquewhite}{rgb}{0.98, 0.92, 0.84}
\definecolor{azure(web)(azuremist)}{rgb}{0.94, 1.0, 1.0}

\title{\LARGE \bf
Reinforcement Learning for Safety Testing: \\Lessons from A Mobile Robot Case Study
}

\author{Tom P. Huck$^{1}$, Martin Kaiser$^{1}$, Constantin Cronrath$^{2}$,\\ Bengt Lennartson$^{2}$, Torsten Kröger, and Tamim Asfour$^{1}$
\thanks{*This work was supported by the German Federal Ministry for Economic Affairs and Climate Action under the project ``SDM4FZI'' (\url{www.sdm4fzi.de}), and by VINNOVA under the ITEA3 ``AITOC'' project. The support is gratefully acknowledged.}
\thanks{$^{1}$Institute of Anthropomatics and Robotics (IAR), Karlsruhe Institute of Technology, Germany. Corresponding Author:
{\tt\footnotesize tom.huck@kit.edu}}%
\thanks{$^{2}$Division of Systems and Control, Department of Electrical Engineering, Chalmers University of Technology, Gothenburg, Sweden.}%
}

\begin{document}

\maketitle
\thispagestyle{empty}
\pagestyle{empty}

\begin{abstract}
Safety-critical robot systems need thorough testing to expose design flaws and software bugs which could endanger humans. Testing in simulation is becoming increasingly popular, as it can be applied early in the development process and does not endanger any real-world operators. However, not all safety-critical flaws become immediately observable in simulation. Some may only become observable under certain critical conditions. If these conditions are not covered, safety flaws may remain undetected. Creating critical tests is therefore crucial. In recent years, there has been a trend towards using Reinforcement Learning (RL) for this purpose. Guided by domain-specific reward functions, RL algorithms are used to learn critical test strategies.
This paper presents a case study in which the collision avoidance behavior of a mobile robot is subjected to RL-based testing. The study confirms prior research which shows that RL can be an effective testing tool. However, the study also highlights certain challenges associated with RL-based testing, namely (i) a possible lack of diversity in test conditions and (ii) the phenomenon of \textit{reward hacking} where the RL agent behaves in undesired ways due to a misalignment of reward and test specification. The challenges are illustrated with data and examples from the experiments, and possible mitigation strategies are discussed.
\end{abstract}
\section{INTRODUCTION}
\label{sec:introduction}
Robots that interact with or operate in the vicinity of humans require thorough safety analyses prior to commissioning in order to expose safety-critical flaws, such as software bugs or design errors. One approach for such safety analyses is simulation-based testing, which is especially useful in early development stages where physical prototypes are unavailable~\cite{kapinski2016simulation}. 

Simulation-based testing can be computationally expensive. Therefore, it is rarely feasible to exhaustively simulate all possible test cases. Thus, a crucial challenge in simulation-based testing is the creation and selection of safety-critical test cases which expose existing safety flaws. This is difficult since the conditions that lead to safety violations are generally not known a priori. If the tests are not critical enough, potentially hazardous safety flaws could remain undetected in the system~\cite{bolbot2019vulnerabilities}.
\begin{figure}[!h]
    \centering
    \includegraphics[width=1\linewidth]{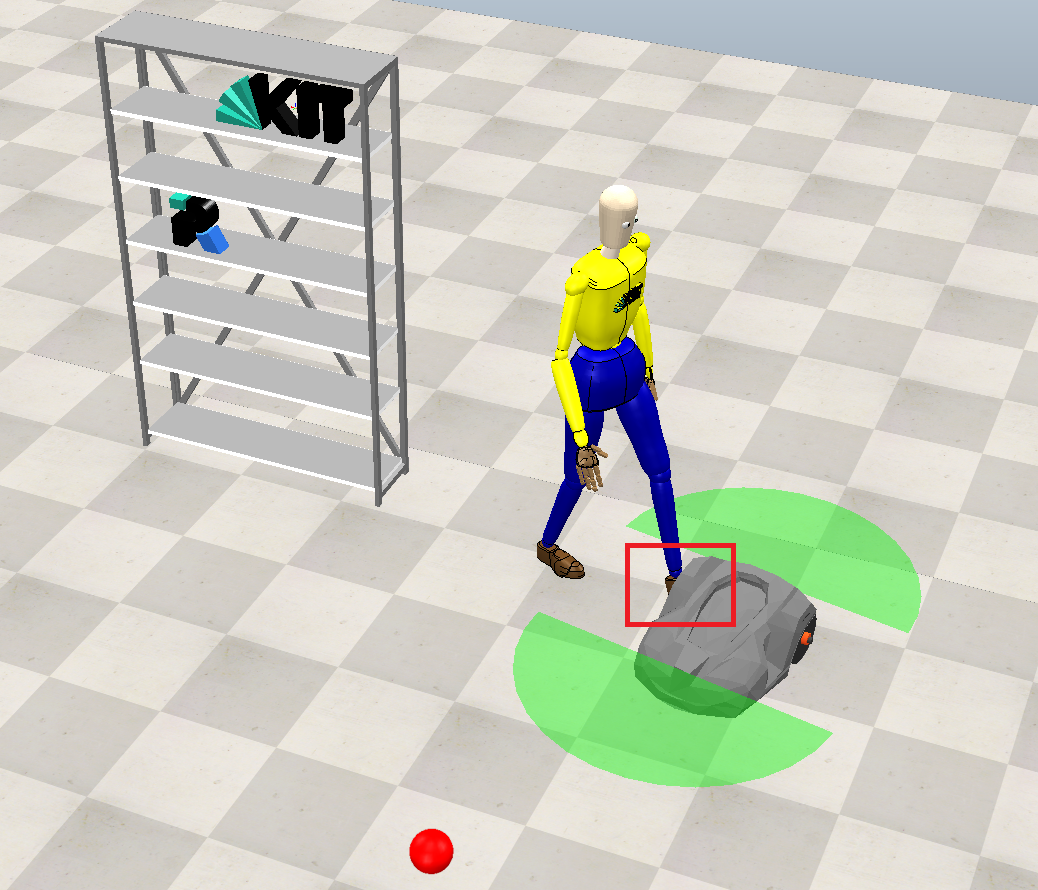}
    \caption{Example of a collision discovered by RL-based testing. Here, the RL agent has learned to exploit a safety flaw (i.e., the gap between the laser scanner fields) to provoke a collision.}
    \label{fig:Simulation}
\end{figure}
One way to address these challenges is the use of Reinforcement Learning (RL) for creating test cases. In RL-based testing, RL algorithms are guided by a reward signal which reflects the criticality of test cases. By attempting to maximize rewards, these algorithms learn critical test strategies. 
While RL-based testing has been demonstrated to be effective in a variety of applications~\cite{yamagata2020falsification,Lee2020}, it is also associated with some challenges and drawbacks which so far have been rarely discussed in literature.

This paper presents lessons learned from a case study where RL-based testing is used to validate the collision avoidance behavior of a mobile robot. In our study, we embed the control code of a real-world controller into a simulation environment and employ RL to create critical test conditions which lead to human-robot collisions (cf. Fig. \ref{fig:Simulation}). We evaluate different variants of RL-based testing and compare it to a baseline random sampling method.
The contributions of this paper are twofold:
\begin{itemize}
    \item[(i)] We demonstrate RL-based testing in a practical use case and describe the implementation of our experiments as guiding examples for similar applications. 
    \item[(ii)] We highlight two critical weak points of RL-based testing, illustrate them with data from our experiments, and discuss mitigation measures. 
\end{itemize}
While our findings are only empirical and limited to a single system under test, we believe that they highlight important caveats that should be considered in practical applications.

\section{RELATED WORK}
\label{sec:related_work}

\subsection{Analysis of Safety Critical Systems}
International safety standards like IEC 61508 \cite{STD_IEC61508} or ISO 12100 \cite{STD_ISO2011} require safety-critical systems to be analyzed thoroughly before commissioning, especially with respect to potential hazards posed to humans. This analysis is often based on human reasoning, expert knowledge, and simple tools such as checklists \cite{Hornung2021,RA_Huck2021b}. Analysts may also use structured analysis methods like \textit{Hazard and Operability Analysis} (HAZOP)~\cite{STD_IEC61882}, \textit{HAZOP-UML}~\cite{Guiochet2016} or \textit{Systems-Theoretic Process Analysis} (STPA) \cite{RA_Leveson2012} which are typically based on graphical system models (e.g., flow diagrams, control structure diagrams, UML-diagrams) and guide words. 

In addition, \textit{rule-based expert systems} can provide automated hazard analyses using domain-specific knowledge encoded in rule-bases. Awad et al. describe such an expert system for safety assessment of HRC applications \cite{RA_Awad2017}. There are also \textit{hybrid approaches} that combine pre-defined rules with simulation environments. Examples in a robotics context are the tools \textit{CobotPlaner} \cite{FraunhoferIFF} and \textit{DynaRisk} \cite{RA_Bdiwi2022}.

\textit{Formal methods} such as model checking can be applied in cases where rigorous safety proofs are required \cite{BaiKat:08,MISC_Clarke2018}. While model checking is typically used for checking software and hardware designs, it can also be leveraged to identify hazards in cyber-physical systems. A prominent tool for this purposes is \textit{SAFER-HRC}, a model checking tool which supports safety assessment of HRC applications \cite{Askarpour2016,Askarpour2017,RA_Askarpour2020}. Furthermore, Rathmair et al. describe a model checking application for safe robotics which is based on the model checker \textit{nuXmv} \cite{RA_Rathmair2021}. 

\subsection{Simulation-based Testing}
For systems with a high degree of complexity, or systems with black-box components, applicability of the aforementioned methods is limited. Methods based on human reasoning do not scale well with increasing complexity, and formal methods run into state-space explosion problems. Furthermore, the presence of black box components inhibits formal analysis. As an alternative, simulation-based testing is often used, for instance in autonomous vehicle testing \cite{Norden2019,Chance2019}, robotics \cite{RA_Araiza2016,RA_Bobka2016a,RA_Huck2021} and aerospace \cite{Lee2015} applications. 

A crucial issue in simulation-based testing is how to create safety-critical test cases, since certain hazards may be overlooked if the simulated scenarios are not sufficiently critical to expose them. This challenge can be addressed by using optimization or RL algorithms to create critical test scenarios \cite{Corso2020Survey,Ding2020}. While this testing method is frequently used for autonomous vehicles \cite{corso2019adaptive}, it is rarely used in the domain of HRC. So far, the approach has been applied by Araiza-Illan et al. \cite{RA_Araiza2016} who use RL to create HRC test cases which maximize code coverage criteria, and in our own prior work \cite{RA_Huck2021}, where we use Monte Carlo Tree Search (MCTS) to create critical behaviors of virtual humans to expose hazards in HRC systems. However, none of the aforementioned works have discussed the issues and challenges which were explored in this case study.

\section{PRELIMINARIES}
This section introduces some preliminaries which are required for the understanding of the following case study.

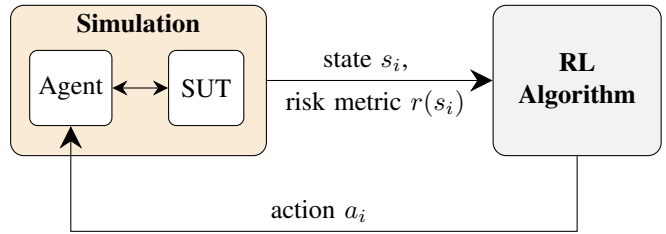
\begin{figure}
    \begin{center}
        \begin{tikzpicture}
          \node[draw, rectangle, minimum width=3.4cm, minimum height=2cm, anchor = south, rounded corners=5pt, fill=antiquewhite] (box1) {};
          \node[below] at ([yshift=0cm]box1.north) {\textbf{Simulation}};
          
          \node[draw, rectangle, minimum width=1cm, minimum height=1cm, rounded corners=3pt, fill=white, anchor=center] at ([xshift=-0.9cm,yshift=-0.1cm]box1.center) (innerbox1) {Agent};
          \node[draw, rectangle, minimum width=1cm, minimum height=1cm, rounded corners=3pt, fill=white, anchor=center] at ([xshift=0.9cm,yshift=-0.1cm]box1.center) (innerbox2) {SUT};
        
          \draw[{Stealth[length=2mm, width=2mm]}-{Stealth[length=2mm, width=2mm]}] (innerbox1.east) -- node[above] {} (innerbox2.west);
        
          \node[draw, rectangle, minimum width=2cm,  text width=2cm, minimum height=2cm, rounded corners=5pt, fill=anti-flashwhite, right=3cm of box1,align=center] (box2) {\textbf{RL Algorithm}};
        
          \draw[-{Stealth[length=3mm, width=3mm]}] (box1.east) -- node[xshift=0.5cm, above, text width=2.5cm] {state $s_i$,} node[below, text width=2.5cm] {risk metric $r(s_i)$} (box2.west);
        
          \draw[-{Stealth[length=3mm, width=3mm]}] (box2.south) -- ++(0,-1cm) -| (innerbox1.south) node[xshift=3.3cm, midway, above] {action $a_i$};

        \end{tikzpicture}
    \caption{RL-based testing: The RL algorithm selects an action which is performed by the agent in simulation environment, jointly with the SUT. The resulting behavior is evaluated w.r.t. a risk metric that is fed back.}
    \label{fig:rl_based_testing}
    \end{center}
\end{figure}

\subsection{Reinforcement Learning}
Reinforcement learning can be considered a form of direct adaptive optimal control, in which the reinforcement learning agent estimates an optimal policy directly from system interactions~\cite{sutton1992reinforcement}. More formally, the goal of the RL agent is to maximize the expected sum of discounted future rewards, i.e.:
\begin{align}
     & \text{maximize} \quad \mathbb{E}_\pi \left[ \sum_{k=0}^{\infty} \gamma^k r_{k} \right] \label{eq:discounted_rewards} \\
     & \text{subject to} \quad s_{k+1} = \mathcal{M}(s_k, a_k) \quad \forall k \in [0,\infty)~,
\end{align}
by learning a control policy $\pi(s): S \mapsto A$ that fulfills this goal. Here, $S$ is the state space, $A$ is the action space, $k$ is the index of the discrete time step, $\gamma$ is a discount factor, $r_{k}$ is the reward received in time step $k$, and $\mathcal{M}$ are the dynamics of the system. RL algorithms can be primarily categorized into model-based (such as Dynamic Programming) and model-free methods that learn purely based on data sampled from the system. An exact (tabular) and an approximate (sometimes `deep') version exist for most algorithms. Model-free methods can be further divided into temporal-difference (TD-learning) methods that exploit estimation errors, and policy optimization methods that tune the parameters of the policy directly. Actor-Critic methods combine policy optimization and TD-learning. The advantage of model-free deep reinforcement learning is its ability to learn reward-maximizing control policies for unknown systems and arbitrary reward functions. It is therefore suitable for automatically uncovering unexpected safety hazards. 

\subsection{RL-based Safety Testing}
The principle of RL-based safety testing is illustrated in Fig.~\ref{fig:rl_based_testing}: The RL algorithm chooses an action which is performed by the agent in the simulation and causes the system under test (SUT) to react. After executing the action, the agent receives a new state along with a reward that guides the agent towards creating more critical behaviors.

Thus, an RL-based safety testing problem can be characterized by the following tuple:
\begin{equation}
    \langle S, A, \mathcal{M}, risk, s_0, \Phi \rangle
\end{equation}
Where the dynamics model $\mathcal{M}$ is given by a simulation of a SUT and an RL agent, $S$ is the state-space of the simulation, $A$ is the action space of the agent, and $risk$ is a risk metric which maps to each simulation state a scalar non-negative value that quantifies the level of risk which is associated with that state:
\begin{equation}
risk: S \rightarrow \mathbb{R}^+_0
\end{equation}
The risk metric is used to calculate the reward $r$ for the agent. Further, $s_0\in S$ is the initial state of the simulation and $\Phi$ is a safety specification, that is, a user-defined function which maps each simulation state to a binary safety decision (i.e., safe or unsafe).
Assuming that the SUT reacts deterministically to a given agent behavior, the evolution of the simulation state is determined by the agent's choice of actions. When executed in the simulator, a given sequence of actions $(a_1, a_2, \dots, a_n),\quad a_i \in A$ will therefore result in a sequence of simulation states $(s_0,s_1,\dots, s_m),\quad s_i \in S$. Since the overall goal of the testing problem is to find unsafe states, the agent needs to find an action sequence for which the resulting state trajectory contains at least one unsafe state according to $\Phi$.

\subsection{Reward Hacking}
\label{sec:reward_hacking}
Despite the promises of RL-based testing, formulating the testing problem remains challenging. The test agent would ideally receive a continuous stream of rewards that guide it quickly towards high-risk situations. However, seemingly helpful or innocent aspects of the test formulation may result in undesired, but provably `optimal' behavioral policies. Such an undesired discrepancy between a learned behavior and the behavior that the designer of the test intended to invoke or finds acceptable, is often called \textit{reward hacking} \cite{amodei2016concrete}. In the context of simulation-based testing, this may have two causes: 1) simplifications or bugs in the simulator or, 2) a discrepancy between the implemented and truly intended reward function. Often, the effect of a particular reward function design on the optimal policy is not immediately obvious, because an optimal policy is obtained from maximizing the expected sum of discounted future rewards generated by the policy and simulator dynamics, while a reward function must be designed in the state action space. This may lead to testing agents that either `farm' intermediate rewards, or prematurely force terminal rewards. Special care must thus be exercised in the design of the testing problem to minimize the risk for reward hacking.


\section{CASE STUDY} \label{sec:method}

\subsection{Use Case and Setup}
\begin{figure}
    \centering
    \includegraphics[width=1\linewidth]{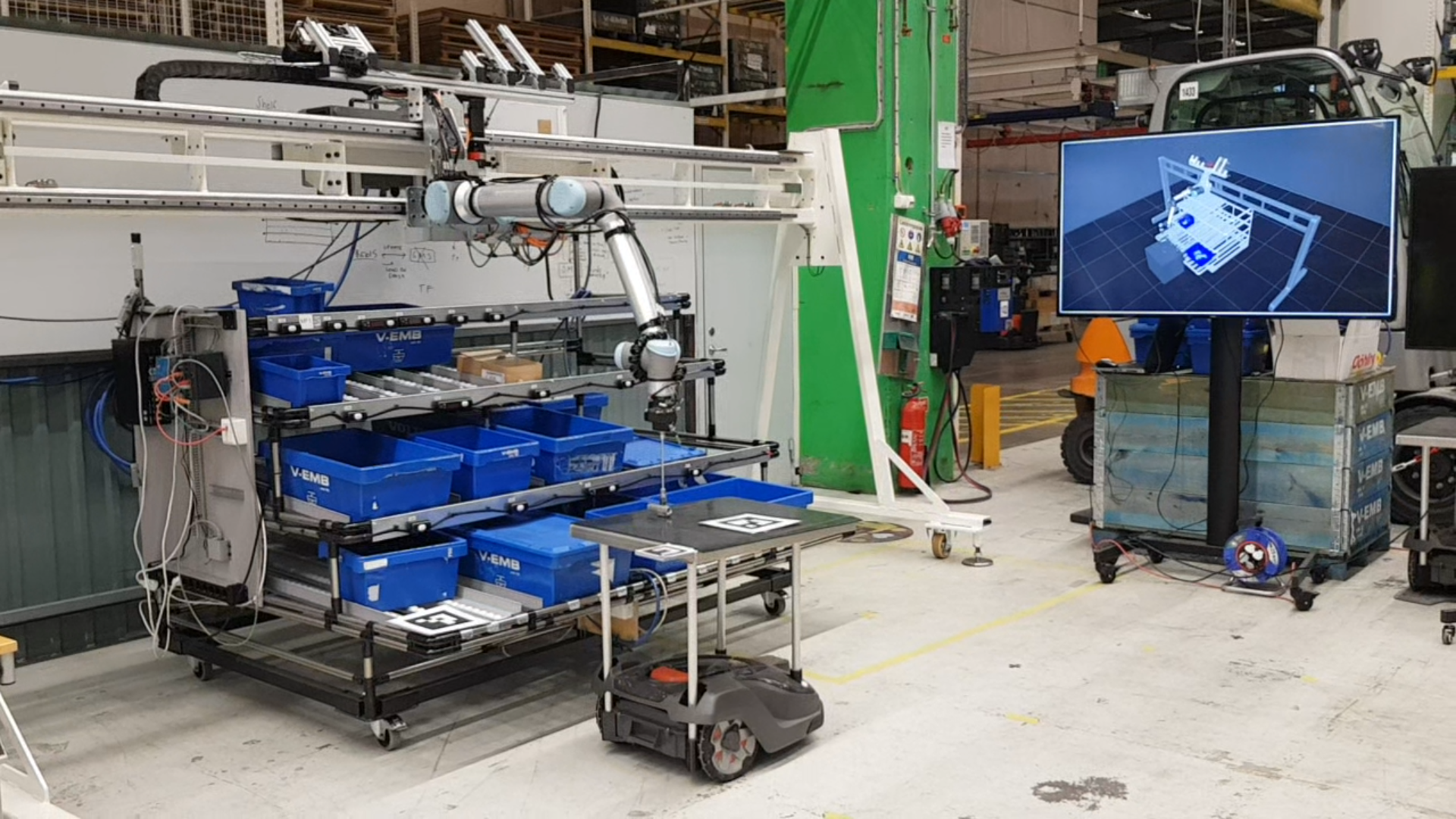}
    \caption{Industrial application: A mobile robot being loaded with parts for delivery on the factory shopfloor. The robot needs to navigate autonomously to deliver its payload.}
    \label{fig:use_case}
\end{figure}

In the given case study, a mobile robot is subjected to RL-based testing. This application is motivated by a practical use case in automotive manufacturing (Fig.~\ref{fig:use_case}). The mobile robot's task involves transporting specific parts to their designated locations on the shop floor. As human workers are present on the shop floor, it is essential for the robot's controller to avoid collisions. Ceiling-mounted cameras track the position of robot and worker and feed this information to the controller. In addition, the controller has a map of static obstacles. The robot is controlled by a nonlinear model predictive controller (NMPC), which can predict both the human's future movement as well as the robot's dynamics. As a backup safety measure, the robot also has laser scanner fields (green areas in Fig.~\ref{fig:Simulation}) which trigger safety stops if they detect obstacles, thereby overriding the controller.

Mobile robot and human are modeled in the simulator \textit{CoppeliaSim}. The NMPC controller is connected to the simulation environment via a custom interface. At each simulation timestep, the NMPC is provided with the present position and heading of the human. The NMPC calculates a prediction of the human's movement and generates an appropriate control action for the robot, which is subsequently executed within the simulation. A description of NMPC is omitted for brevity. We refer to \cite{Berlin2021} instead. The resulting trajectories are logged along with other safety-relevant information.

In our setup, the human model acts as agent while the mobile robot is the SUT. Hence, the objective of the reinforcement learning task is to identify critical human trajectories that lead to collisions with the robot. At any step~$k$, the agent chooses an action 
\begin{equation}
    a_k=\left(v_k,\dot{\theta}_k\right) 
\end{equation}
where $v_k \in [0,1.5]$ denotes the walking speed of the human 
model in $\frac{m}{s}$ and $\dot{\theta}_k \in [-\pi/2, \pi/2]$ denotes the turning rate in ${rad}/{s}$. 

We specify that any collision between human and robot is considered an unsafe state, with two exceptions, namely (i) cases where the human walks into the robot while the robot is standing still, and (ii) cases where the human runs into the moving robot from behind, as these two cases do not reflect safety flaws on the robot's side.

\subsection{Risk Metric and Reward Definition}
The reward $r_k$ for the RL algorithm is issued on the basis of the risk metric $risk$ which is calculated at every timestep~$k$:
\begin{equation}
    risk(k) =
    \begin{cases}
            100\cdot \left(1+\frac{v_H(k)\cdot v_R(k)}{v_{max,H}\cdot v_{max,R}}\right)\quad &\text{in case (a)} \\
            e^{-d_{HR}(k)}\quad &\text{in case (b)} \\
            0\quad &\text{otherwise}
    \end{cases}
    \label{eq:risk}
\end{equation}
where $v_H(k)$ and $v_R(k)$ denote the current velocity and $v_{max,H}$ and $v_{max,R}$ denote the maximum velocity of human and robot, respectively. Further, $d_{HR}(k)$ denotes the current human-robot distance. Case (a) applies to collisions states. Here, the risk score takes large values between 100 and 200. Case (b) applies if human and robot are not in contact, but the robot is moving. In this case, a small risk between 0 and 1 is given, which decreases exponentially with the distance. In all other cases, the risk metric remains zero. We emphasize that the risk metric chosen here is a simplified heuristic to guide the RL algorithms. Other forms of risk representation are also possible. On the basis of this risk metric, we employ two different reward formulations:

\textit{Accumulated reward:} At any given timestep, the current value of the risk metric is issued as reward:
\begin{equation}
    r_k = risk(k)
    \label{eq:r_acc}
\end{equation}
This results in an accumulation of risk scores over an action sequence episode (cf. Eq. (\ref{eq:discounted_rewards})).

\textit{Maximum reward}: During an action sequence, the risk score is calculated and stored, but the issued reward is zero in all timesteps but the last. In the $n$-th (i.e., last) timestep, the maximum of the previously stored risk scores values is issued as a reward:
\begin{equation}
    r_k =
    \begin{cases}
        0 &k=1..(n-1)\\
        \max_{\tilde{k}=1}^{n} risk\left(\tilde{k}\right) &k=n
    \end{cases}
    \label{eq:r_max}
\end{equation}
Thus, the agent is rewarded only for the single highest risk score that has occurred over the course of an action sequence, and not for accumulation of risk scores.

\subsection{Algorithms}
We selected the Proximal Policy Optimization algorithm (PPO) as an RL algorithm due to its relative ease of use and the availability of a well-tested state-of-the-art implementation by OpenAI\footnote{\footnotesize \url{https://openai.com/research/openai-baselines-ppo}}. For an explanation of the algorithm we refer to \cite{schulman2017proximal}. The algorithm is deployed with both the accumulated and the maximum reward formulation (cf. Eq. (\ref{eq:r_acc})-(\ref{eq:r_max}))

As baseline, we deploy a random sampling approach where actions are sampled from a uniform distribution over the agent's action space. However, we found that pure random sampling frequently leads to erroneous behaviors (e.g., walking into walls or other static obstacles), resulting in weak baseline performance. For a fairer comparison, we therefore augment the baseline with a simple collision avoidance scheme which restricts the set of feasible actions to a subset that does not lead to collision with static obstacles within the next timestep.

Henceforth, the RL algorithms with accumulated and maximum rewards will be abbreviated by $RL_{acc}$ and $RL_{max}$, respectively, and the random sampling baseline by $RS$.

\subsection{Experiments}
\begin{figure*}
    \centering
        \subfloat{\includegraphics[width=0.47\textwidth]{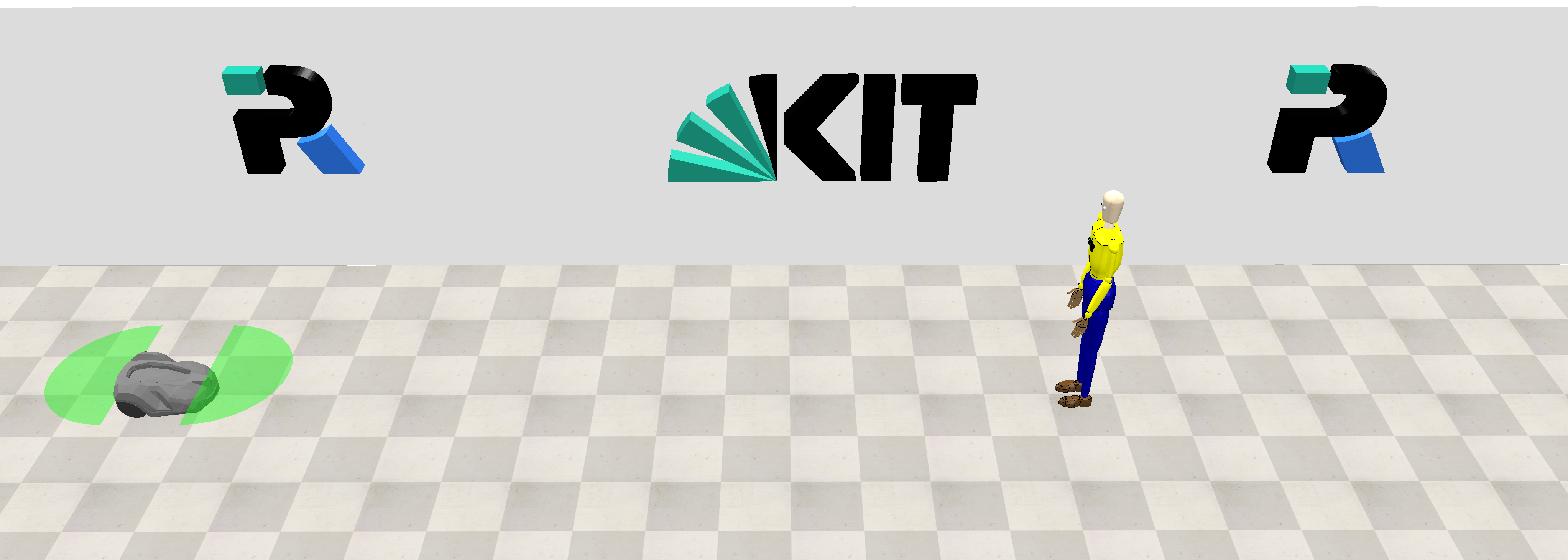}}
    \hfill
        \subfloat{\includegraphics[width=0.47\textwidth]{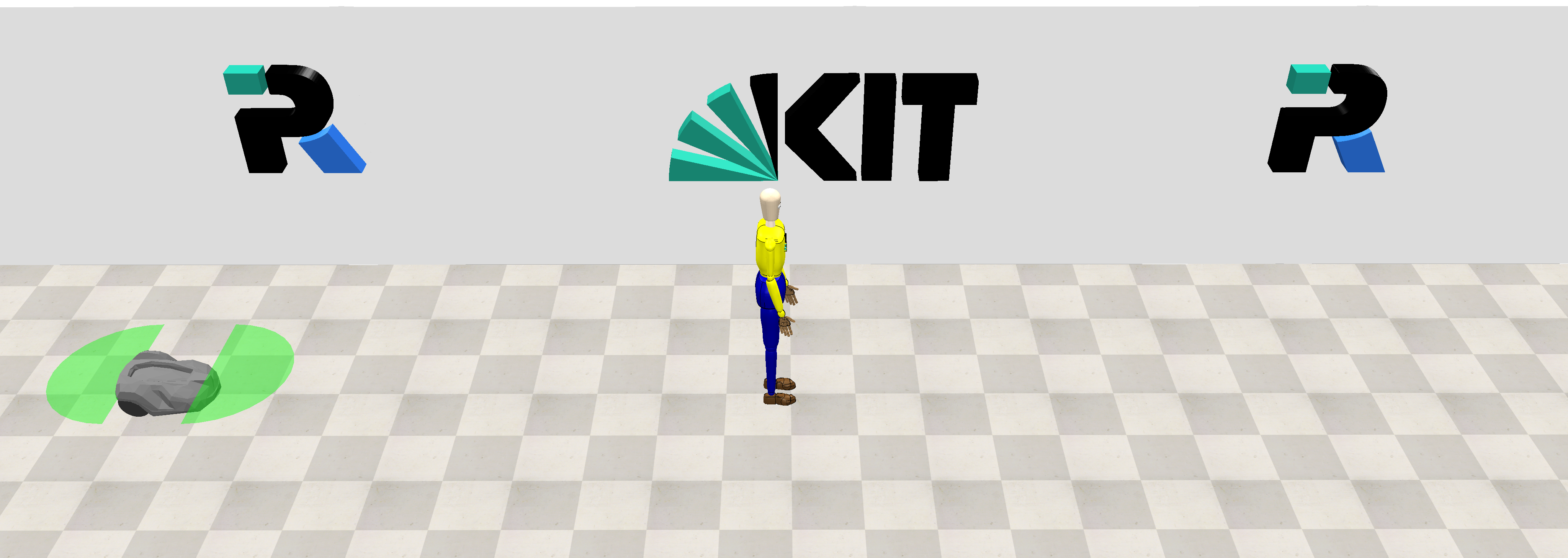}}
    \caption{Initial states: In the \textit{head-on} state (left), agent and AGV initially face each other. In the \textit{overtaking} state (right), both face the same direction (right). The \textit{crossing} initial state, where human and AGV start perpendicular to each other, is depichted in Fig. \ref{fig:Simulation}}
    \label{fig:scenarios}
\end{figure*}

We perform experiments starting from three different initial simulation states:
\begin{itemize}
    \item \textit{Crossing:} human and robot start in a perpendicular configuration (Fig. \ref{fig:Simulation})
    \item \textit{Head-on:} human and robot start opposite to each other (Fig. \ref{fig:scenarios}, left)
    \item \textit{Overtaking:} the robot starts behind the human, both facing the same direction  (Fig. \ref{fig:scenarios}, right)
\end{itemize}
Furthermore, we consider three different SUT settings: 
\begin{itemize}
    \item $L_{none}$: the SUT is only controlled by the NMPC without an additional laser scanner.
    \item $L_{1.25m}$: the laser scanner field has a radius of 1.25m.
    \item $L_{2m}$: the large laser scanner field has a radius of 2m.
\end{itemize}

With three initial conditions, three SUT settings, and three algorithms ($RL_{acc}$, $RL_{max}$ and $RS$), there are 27 total configurations in which the simulation can be run. For each of these configurations, three test runs with different random seeds are performed, resulting in 81 test runs overall. Each test run is limited to 20'000 action sequences. Each action sequence starts at the respective initial state and terminates if (i) the mobile robot reaches its goal position, (ii) the safety specification is violated (i.e., a critical collision occurs), or (iii) the duration exceeds 30s in simulation time.

\section{FINDINGS FROM THE CASE STUDY}
Due to the extensive amount of data generated by the experiments, we present the results in an aggregated manner, namely in the form of three lessons that were drawn from the case study. 

\begin{figure}[h!]
    \centering
    \includegraphics[width=1\linewidth]{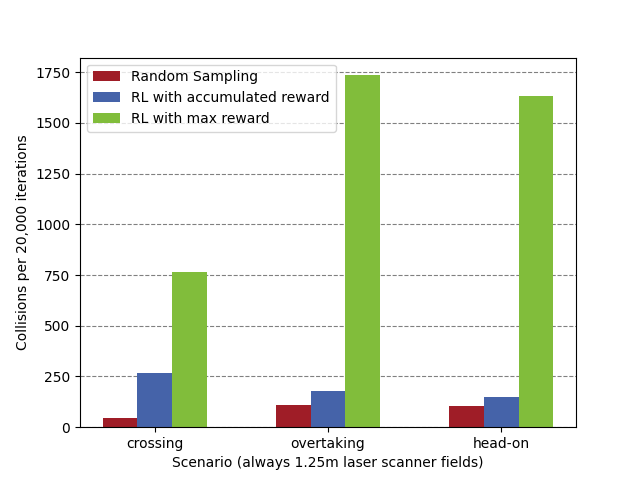}
    \caption{Average number of collisions by test scenario and algorithm.}
    \label{fig:n_collisions}
\end{figure}

\begin{tcolorbox}[colback=aliceblue]
	\textbf{Lesson 1:} RL can be an effective tool to generate critical test cases. In our experiments, it outperformed random sampling in the majority of test cases.
\end{tcolorbox}
To assess the algorithms' effectiveness in finding unsafe states, we evaluated the number of collision states found by both RL variants and the baseline algorithm, respectively. The results indicate that both $RL_{acc}$ and $RL_{max}$ achieve a significantly higher number of collisions than random sampling in the majority of the test configurations. More specifically, across all 27 test configurations, $RL_{max}$ outperformed $RS$ in 24 configurations and $RL_{acc}$ outperformed $RS$ in 19 configurations.
For reasons of brevity, we do not provide a plot for each scenario and configuration here. As an example, Fig.~\ref{fig:n_collisions} shows the results for $S_{1.25m}$. 

\begin{tcolorbox}[colback=aliceblue]
	\textbf{Lesson 2:} Although RL tends to converge to certain unsafe behaviors, the overall diversity of identified unsafe behaviors is on par with random exploration.
\end{tcolorbox}

Above we discussed the number of collisions created by the different algorithms. However, this metric only paints a partial picture of the  effectiveness in finding hazards. It is also important how \textit{diverse} the results are. Especially with RL, there is a possibility that the algorithm may prematurely converge to a specific unsafe behavior which yields a high reward and thereby discover the same hazardous situation over and over while others remain undiscovered. To avoid this, the identified unsafe states should be as diverse as possible to expose a wide range of potential hazards. In order to quantify this diversity, we classified the identified collisions according to a number of domain-specific criteria such as collision direction, velocity, and control mode of the AGV at time of collision (e.g., if the AGV's laser scanner had been triggered or not). We found that according to our set of criteria, there were nine possible types of collisions (eight for the configurations without laser scanner). We then evaluated how many of these collision types were found by the respective algorithms.

The results are shown in Table \ref{tab:collision_classes_found}. $RL_{max}$ outperforms $RS$ in four cases while being outperformed by $RS$ only once (scenario \textit{overtaking}, $S_{2m}$). $RL_{acc}$ performs slightly worse than $RL_{max}$, but still slightly better than $RS$ (i.e., outperforms $RS$ in two cases while being outperformed once). Although $RL_{acc}$ and $RL_{max}$ might not exhibit a substantial advantage over $RS$ in this aspect, it can be asserted that the anticipated drawback stemming from early convergence did not actually occur.
Furthermore, it can be argued that while RL in our experiments did not necessarily find more collisions classes, it found a higher number of collisions \textit{per class}, thereby leading to a more reliable detection of hazards.

\begin{table}
    \centering
    \caption{Collision classes found per scenario and configuration.}
    \label{tab:collision_classes_found}
    \normalsize
    \begin{tabular}{l l c c c} 
        &  & $RL_{max}$ & $RL_{acc}$ & RS \\
        \midrule
        \textit{crossing} & $L_{none}$ & 8\,/\,8 & 8\,/\,8 & 8\,/\,8 \\
                          & $L_{1.25m}$ & 6\,/\,9 & 7\,/\,9 & 5\,/\,9 \\
                          & $L_{2m}$ & 8\,/\,9 & 6\,/\,9 & 7\,/\,9 \\
        \midrule        
        \textit{head-on} & $L_{none}$ & 8\,/\,8 & 8\,/\,8 & 8\,/\,8 \\
                         & $L_{1.25m}$ & 5\,/\,9 & 5\,/\,9 & 5\,/\,9 \\
                         & $L_{2m}$ & 7\,/\,9 & 6\,/\,9 & 6\,/\,9 \\
        \midrule        
        \textit{overtaking} & $L_{none}$ & 8\,/\,8 & 8\,/\,8 & 8\,/\,8 \\
                            & $L_{1.25m}$ & 7\,/\,9 & 6\,/\,9 & 5\,/\,9 \\
                            & $L_{2m}$ & 5\,/\,9 & 7\,/\,9 & 7\,/\,9 
    \end{tabular}
\end{table}

\begin{figure*}
    \subfloat[Worker walks in front of AGV \label{subfig:reward_hacking_1}]{
        \includegraphics[width=0.24\textwidth]{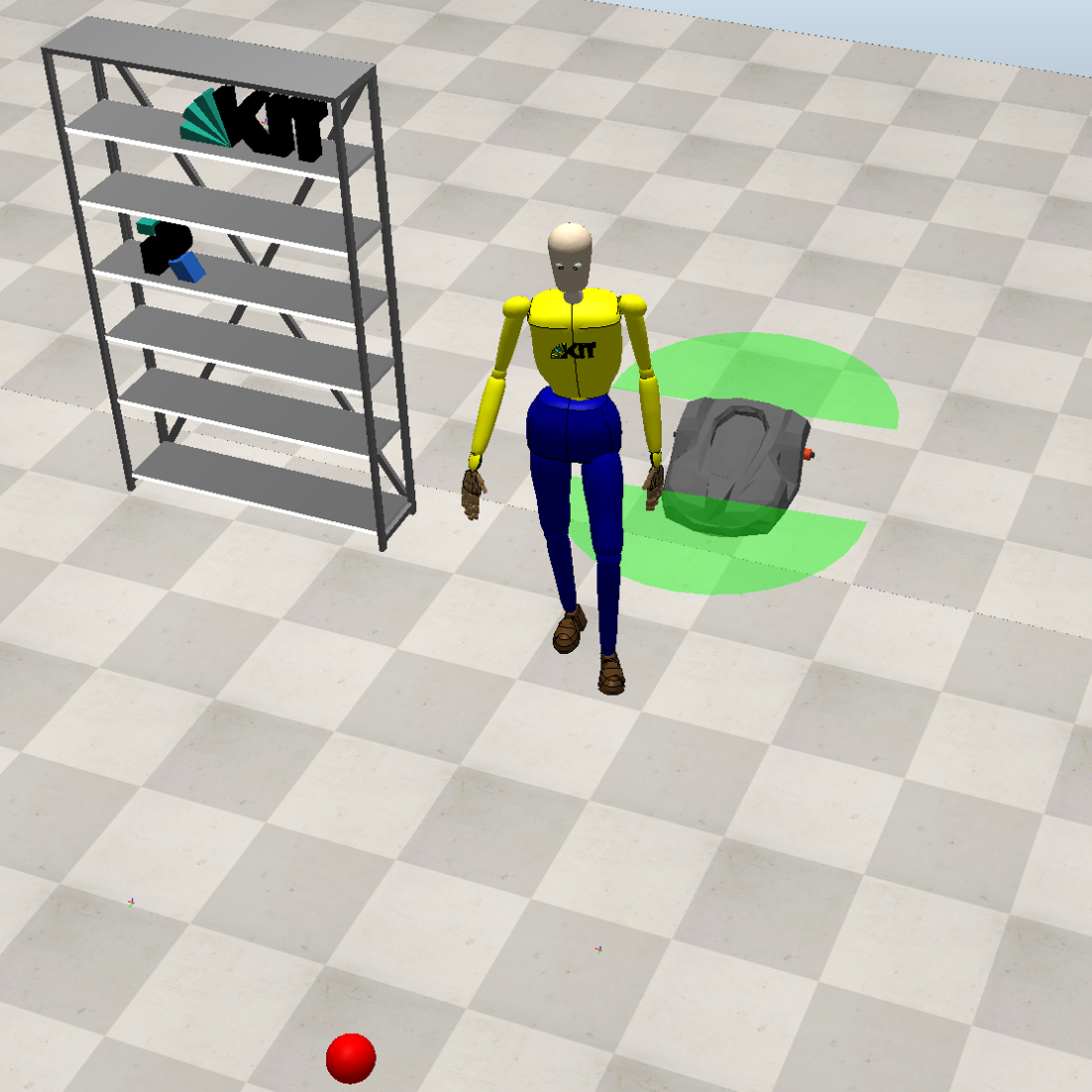}
    }
    \subfloat[Laser scanner field triggers emergency stop \label{subfig:reward_hacking_2}]{
        \includegraphics[width=0.24\textwidth]{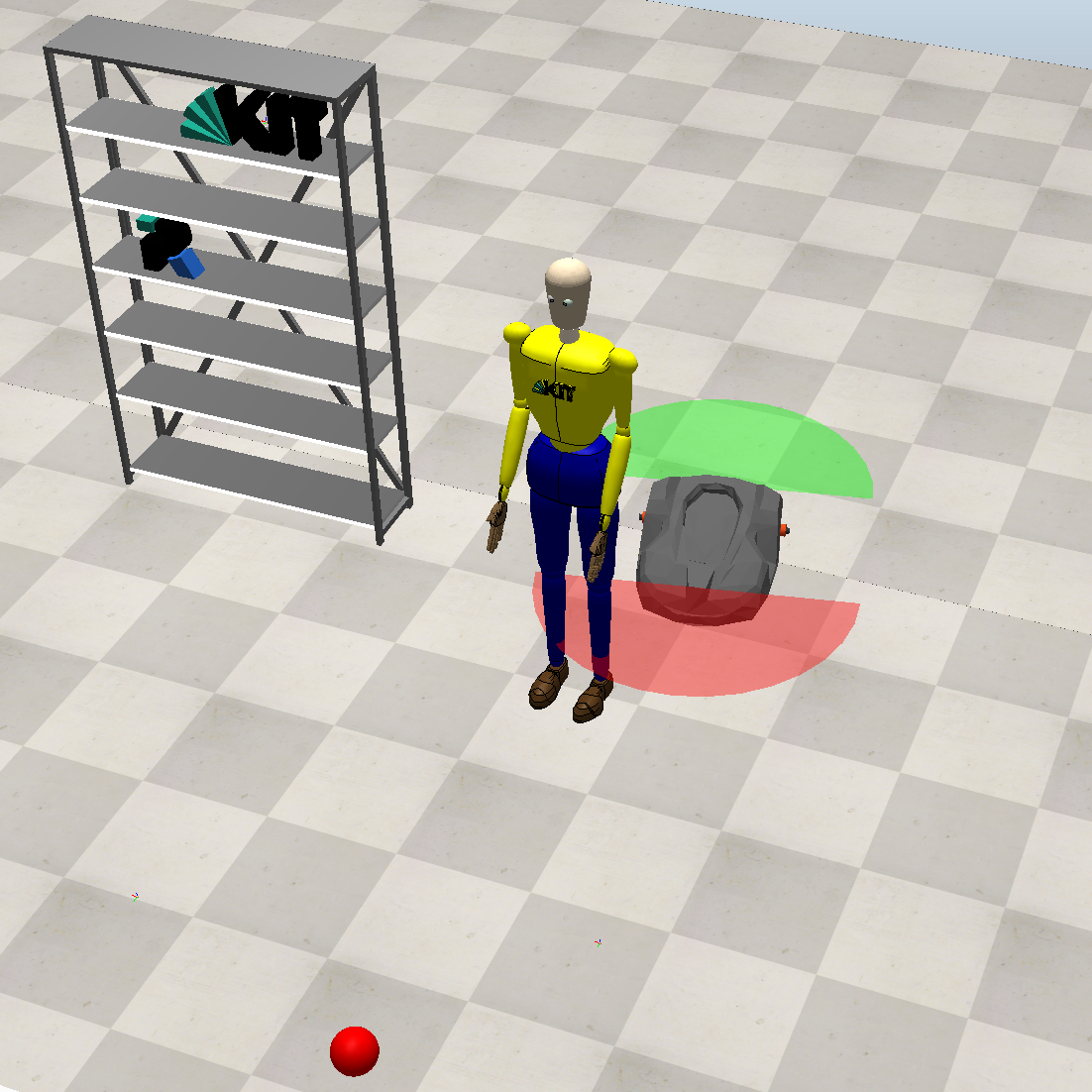}
    }
    \subfloat[Worker leaves laser scanner field \label{subfig:reward_hacking_3}]{
        \includegraphics[width=0.24\textwidth]{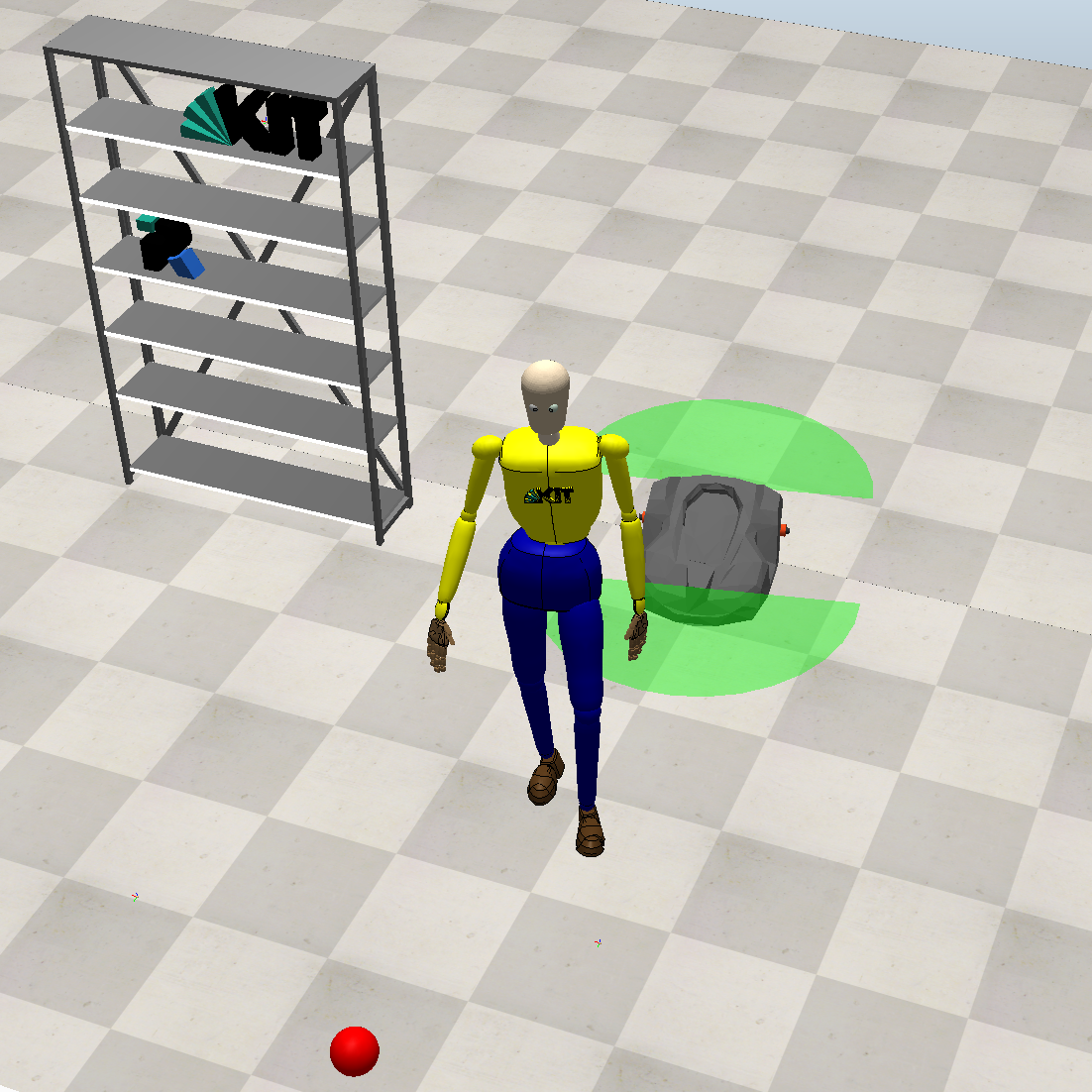}
    }
    \subfloat[Laser scanner field triggers another emergency stop \label{subfig:reward_hacking_4}]{
        \includegraphics[width=0.24\textwidth]{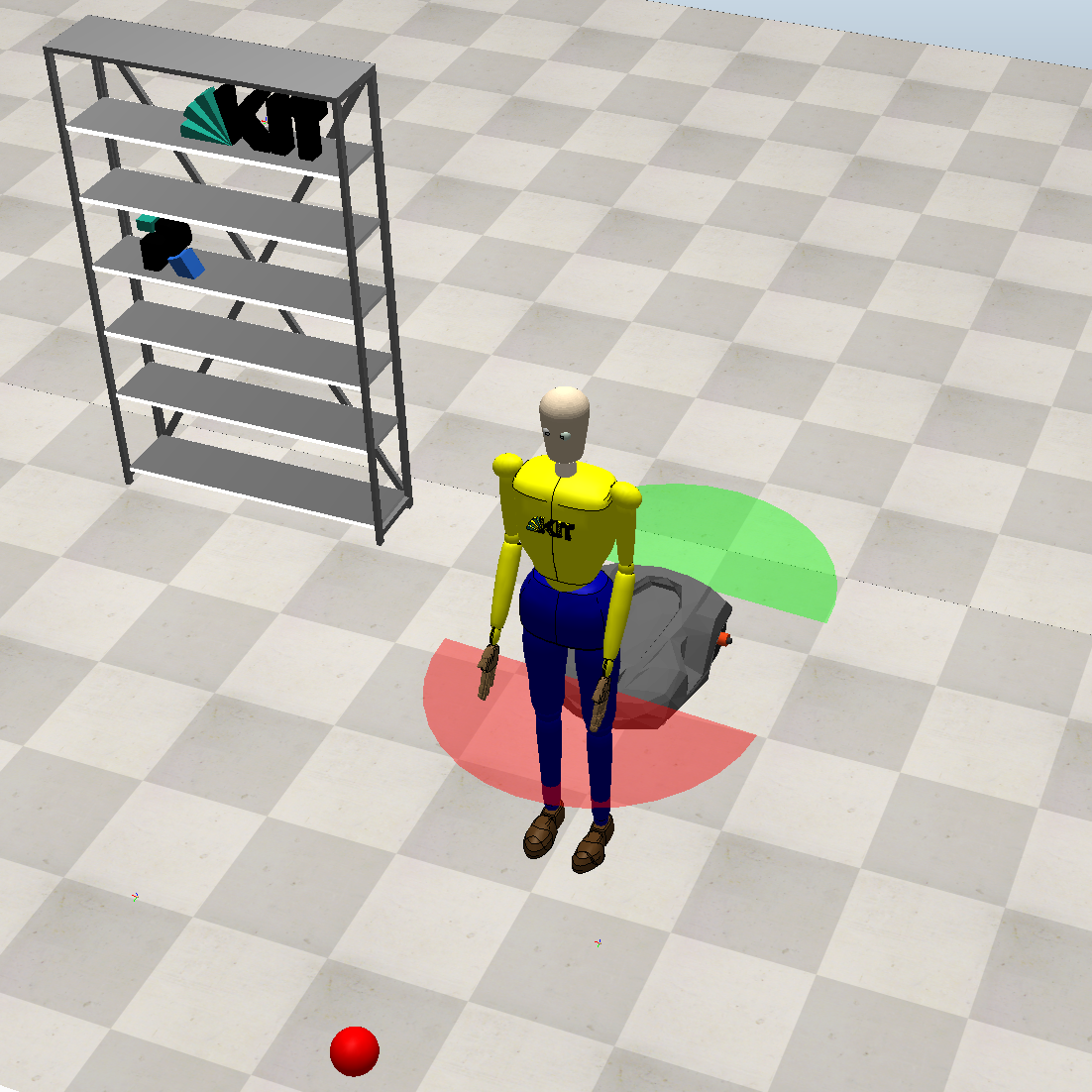}
    }    
    \caption{Example of reward hacking. The AGV tries to reach the red target point.}
    \label{fig:results3}
\end{figure*}

\begin{tcolorbox}[colback=aliceblue]
	\textbf{Lesson 3:} Reward Hacking can severely impede the ability of RL to find unsafe states. However, an appropriate definition of the reward function can mitigate or even solve this issue.
\end{tcolorbox}
When using $RL_{acc}$, the RL agent frequently exhibited behaviors which are to be considered ``reward hacking" according to the characterization in Sec.~\ref{sec:reward_hacking}. Fig.~\ref{fig:results3} shows an example of such a reward hacking behavior: Instead of learning to create a collision, the agent learns to walk in front of the robot, getting close to incur a high reward, and then moving out of the laser scanner zone again to facilitate a re-start of the robot. This behavior is performed repeatedly until the maximum duration of the action sequence in order to incur a high cumulative reward. In some cases, this behavior was so pronounced that it completely stalled the simulations. To counter this issue and enable further experiments, we were forced to introduce an additional termination criterion where the action sequence terminates if the agent is within the laser scanner zone for a prolonged period of time without causing a collision. However, even with these additional termination criteria, the problem was mitigated, but not entirely avoided. With the frequent occurrence of such reward hacking behaviors, the simulations still required significantly more computation time for the same amount of episodes, as shown in Table \ref{tab:runtimes}.

This led us to introduce $RL_{max}$ which uses the maximum reward instead of the accumulated reward. With this reward formulation, the agent has no incentive anymore to accumulate rewards over multiple timesteps by repeatedly creating medium-risk situations. Instead, the agent aims to create maximally risky situations, even if these are only occur in a single timestep. This change alleviated the problem and no occurrences of reward hacking were found in the respective test runs. As seen in Table \ref{tab:runtimes}, this is also confirmed by the significantly lower average computation time. As discussed above, besides the improvement in computation time, this change also led to improvements in the number of collisions and their diversity.

\begin{table*}
    \centering
    \caption{Comparison of the runtimes (in hh:mm) for different reward functions.}
    \label{tab:runtimes}
    \normalsize
    \begin{tabular}{l c c c c c c c c c} 
        & \multicolumn{3}{c}{crossing} & \multicolumn{3}{c}{overtaking} & \multicolumn{3}{c}{head-on}\\
        \cmidrule(r){2-4} \cmidrule(r){5-7} \cmidrule(r){8-10}
        & $L_{1.25m}$ & $L_{2m}$ & $L_{none}$ & $L_{1.25m}$ & $L_{2m}$ & $L_{none}$ & $L_{1.25m}$ & $L_{2m}$ & $L_{none}$\\ 
        \midrule
        avg runtime $RL_{acc}$ & 42:13 & 47:50 & 47:47 & 77:45 & 50:16 & 52:51 & 77:03 & 48:44 & 64:43\\
        avg runtime $RL_{max}$ & 34:23 & 25:34 & 33:49 & 34:57 & 19:41 & 37:24 & 42:47 & 28:28 & 52:40\\
        \midrule
        speedup & 18.6\% & 45.4\% & 29.2\% & 55.0\% & 60.8\% & 29.2\% & 44.5\% & 41.6\% & 18.6\%\\
    \end{tabular}
\end{table*}

\section{SUMMARY AND DISCUSSION} \label{sec:discussion}
The findings from our case study confirm prior studies indicating that RL can be an effective tool for safety testing. However, our case study also highlighted two potential drawbacks.

First, while RL tends to find a larger number of unsafe states, these unsafe states are not necessarily more \textit{diverse}. Instead, RL frequently re-discovers similar unsafe states. Although in our experiments, RL achieved a slightly better diversity than the random baseline, the comparative advantage is minor.

Second, RL-based safety testing is susceptible to the phenomenon of reward hacking. In particular, we noted a tendency for the RL agent to accumulate rewards by performing medium-risk behaviors over a long duration rather than creating high-risk situations. For testing safety-critical systems, the occurrence of a high-risk situation is however more interesting. Therefore, the conventional RL formulation with accumulated rewards may not necessarily be the most suitable formulation for RL-based safety testing. Alternatively, a maximum reward formulation should be considered where appropriate. 

We emphasize that these findings are empirical and restricted to a single case study. Furthermore, our study is limited since it only considers a single SUT, although we aimed to mitigate this limitation by the use of different test configurations. We therefore do not claim that our findings hold generally. Nevertheless, we believe that our findings do highlight important considerations and potential problems that one should take into account when employing RL for testing purposes.  

\section{FUTURE WORK}
Future research should address the aforementioned issues of lacking diversity and reward hacking.

With respect to diversity, an interesting approach is proposed by Corso et al.~\cite{corso2019adaptive}, who use dissimilarity rewards to encourage a higher diversity of generated test cases. This approach could be incorporated into our future studies. Furthermore, the assessment of diversity could be based on clustering algorithms rather than manually defined classes.

Regarding reward hacking, it would be interesting to examine this phenomenon in the context of RL-based testing more systematically. A broader picture of this issue is needed to assess in how far it presents a danger to RL-based testing. Also, reward hacking strongly depends on the nature of the reward formulation. It may therefore be useful to examine whether there are certain classes of testing problems which are particularly susceptible to this phenomenon, and for which certain reward formulations are more suitable than others.

Finally, it should be noted that our study focused on testing the robot controller. As a software component, the controller is affected by systematic errors (e.g., bugs or design flaws). However, hazards can also be caused by random errors and statistical uncertainties (e.g., measurement- or parameter-uncertainties). Exploring how such random effects can be incorporated into simulation-based testing is another important area of future research. In \cite{baek2023uncertainty}, we lay out how this may be achieved.


\bibliography{references.bib}

\begin{thebibliography}{10}
\providecommand{\url}[1]{#1}
\csname url@samestyle\endcsname
\providecommand{\newblock}{\relax}
\providecommand{\bibinfo}[2]{#2}
\providecommand{\BIBentrySTDinterwordspacing}{\spaceskip=0pt\relax}
\providecommand{\BIBentryALTinterwordstretchfactor}{4}
\providecommand{\BIBentryALTinterwordspacing}{\spaceskip=\fontdimen2\font plus
\BIBentryALTinterwordstretchfactor\fontdimen3\font minus
  \fontdimen4\font\relax}
\providecommand{\BIBforeignlanguage}[2]{{%
\expandafter\ifx\csname l@#1\endcsname\relax
\typeout{** WARNING: IEEEtran.bst: No hyphenation pattern has been}%
\typeout{** loaded for the language `#1'. Using the pattern for}%
\typeout{** the default language instead.}%
\else
\language=\csname l@#1\endcsname
\fi
#2}}
\providecommand{\BIBdecl}{\relax}
\BIBdecl

\bibitem{kapinski2016simulation}
J.~Kapinski, J.~V. Deshmukh, X.~Jin, H.~Ito, and K.~Butts, ``Simulation-based
  approaches for verification of embedded control systems: An overview of
  traditional and advanced modeling, testing, and verification techniques,''
  \emph{IEEE Control Systems Magazine}, vol.~36, no.~6, pp. 45--64, 2016.

\bibitem{bolbot2019vulnerabilities}
V.~Bolbot, G.~Theotokatos, L.~M. Bujorianu, E.~Boulougouris, and D.~Vassalos,
  ``Vulnerabilities and safety assurance methods in cyber-physical systems: A
  comprehensive review,'' \emph{Reliability Engineering \& System Safety}, vol.
  182, pp. 179--193, 2019.

\bibitem{yamagata2020falsification}
Y.~Yamagata, S.~Liu, T.~Akazaki, Y.~Duan, and J.~Hao, ``Falsification of
  cyber-physical systems using deep reinforcement learning,'' \emph{IEEE
  Transactions on Software Engineering}, 2020.

\bibitem{Lee2020}
R.~Lee, O.~J. Mengshoel, A.~Saksena, R.~W. Gardner, D.~Genin, J.~Silbermann,
  M.~Owen, and M.~J. Kochenderfer, ``Adaptive stress testing: Finding likely
  failure events with reinforcement learning,'' \emph{Journal of Artificial
  Intelligence Research}, vol.~69, 2020.

\bibitem{STD_IEC61508}
{IEC}, ``{IEC} 61508-1:2010-1 {F}unctional safety of
  electrical/electronic/programmable electronic safety-related systems - {P}art
  1: {G}eneral requirements,'' {I}nternational {E}lectrotechnical {C}ommission,
  2006.

\bibitem{STD_ISO2011}
{ISO}, ``{ISO 12100:2011}: {Safety of machinery - General principles for design
  - Risk assessment and risk reduction},'' 2011.

\bibitem{Hornung2021}
L.~Hornung and C.~Wurll, ``Human-robot collaboration: a survey on the state of
  the art focusing on risk assessment,'' in \emph{Berichte aus der Robotik -
  Robotix-Academy Conference for Industrial Robotics (RACIR) 2021}, Sep. 2021,
  pp. 10--17.

\bibitem{RA_Huck2021b}
T.~P. Huck, N.~M{\"u}nch, L.~Hornung, C.~Ledermann, and C.~Wurll, ``{Risk
  assessment tools for industrial human-robot collaboration: Novel approaches
  and practical needs},'' \emph{Safety Science}, vol. 141, 2021.

\bibitem{STD_IEC61882}
``{IEC} 61882:2016: {H}azard and operability studies ({HAZOP} studies) -
  application guide,'' {I}nternational {E}lectrotechnical {C}ommission, 2016.

\bibitem{Guiochet2016}
J.~Guiochet, ``Hazard analysis of human--robot interactions with hazop--uml,''
  \emph{Safety science}, vol.~84, pp. 225--237, 2016.

\bibitem{RA_Leveson2012}
N.~Leveson, \emph{Engineering a safer world: Systems thinking applied to
  safety}.\hskip 1em plus 0.5em minus 0.4em\relax MIT {P}ress, 2011.

\bibitem{RA_Awad2017}
R.~{Awad}, M.~{Fechter}, and J.~{van Heerden}, ``{Integrated risk assessment
  and safety consideration during design of HRC workplaces},'' in \emph{22nd
  IEEE International Conference on Emerging Technologies and Factory Automation
  (ETFA)}, Sep. 2017.

\bibitem{FraunhoferIFF}
\BIBentryALTinterwordspacing
F.~IFF. (2023) {Cobot Planner - Design Safe HRC Applciations Quickly and
  Easily} (online tool). [Online]. Available:
  \url{https://www.cobotplaner.de/preambel}
\BIBentrySTDinterwordspacing

\bibitem{RA_Bdiwi2022}
M.~Bdiwi, ``{Intuitive Roboterprogrammierung und intelligente Werkzeuge},''
  \emph{JOT Journal f{\"u}r Oberfl{\"a}chentechnik}, vol.~62, no.~8, pp.
  18--19, 2022.

\bibitem{BaiKat:08}
C.~Baier and J.-P. Katoen, \emph{Principles of Model Checking}.\hskip 1em plus
  0.5em minus 0.4em\relax {MIT} Press, 2008.

\bibitem{MISC_Clarke2018}
E.~M. Clarke, T.~A. Henzinger, H.~Veith, R.~Bloem \emph{et~al.}, \emph{Handbook
  of model checking}.\hskip 1em plus 0.5em minus 0.4em\relax Springer, 2018,
  vol.~10.

\bibitem{Askarpour2016}
M.~Askarpour, D.~Mandrioli, M.~Rossi, and F.~Vicentini, ``{SAFER-HRC: Safety
  analysis through formal verification in human-robot collaboration},'' in
  \emph{International Conference on Computer Safety, Reliability, and
  Security}.\hskip 1em plus 0.5em minus 0.4em\relax Springer, 2016, pp.
  283--295.

\bibitem{Askarpour2017}
------, ``Modeling operator behavior in the safety analysis of collaborative
  robotic applications,'' in \emph{International Conference on Computer Safety,
  Reliability, and Security}.\hskip 1em plus 0.5em minus 0.4em\relax Springer,
  2017, pp. 89--104.

\bibitem{RA_Askarpour2020}
M.~Askarpour, M.~Rossi, and O.~Tiryakiler, ``{Co-simulation of human-robot
  collaboration: From temporal logic to 3D simulation},'' in \emph{1st Workshop
  on Agents and Robots for Reliable Engineered Autonomy, AREA 2020}, vol.
  319.\hskip 1em plus 0.5em minus 0.4em\relax Open Publishing Association,
  2020, pp. 1--8.

\bibitem{RA_Rathmair2021}
M.~Rathmair, C.~Luckeneder, T.~Haspl, B.~Reiterer, R.~Hoch, M.~Hofbaur, and
  H.~Kaindl, ``Formal verification of safety properties of collaborative
  robotic applications including variability,'' in \emph{2021 30th IEEE
  International Conference on Robot \& Human Interactive Communication
  (RO-MAN)}.\hskip 1em plus 0.5em minus 0.4em\relax IEEE, 2021, pp. 1283--1288.

\bibitem{Norden2019}
J.~Norden, M.~O'Kelly, and A.~Sinha, ``Efficient black-box assessment of
  autonomous vehicle safety,'' \emph{arXiv preprint arXiv:1912.03618}, 2019.

\bibitem{Chance2019}
G.~Chance, A.~Ghobrial, S.~Lemaignan, T.~Pipe, and K.~Eder, ``An
  agency-directed approach to test generation for simulation-based autonomous
  vehicle verification,'' \emph{arXiv preprint arXiv:1912.05434}, 2019.

\bibitem{RA_Araiza2016}
D.~Araiza-Illan, A.~G. Pipe, and K.~Eder, ``Intelligent agent-based stimulation
  for testing robotic software in human-robot interactions,'' in
  \emph{Proceedings of the 3rd Workshop on Model-Driven Robot Software
  Engineering}, 2016, pp. 9--16.

\bibitem{RA_Bobka2016a}
P.~Bobka, T.~Germann, J.~K. Heyn, R.~Gerbers, F.~Dietrich, and K.~Dröder,
  ``Simulation platform to investigate safe operation of human-robot
  collaboration systems,'' in \emph{6th CIRP Conference on Assembly
  Technologies and Systems (CATS)}, vol.~44, 2016, pp. 187 -- 192.

\bibitem{RA_Huck2021}
T.~Huck, C.~Ledermann, and T.~Kröger, ``Virtual adversarial humans finding
  hazards in robot workplaces,'' in \emph{2021 IEEE International Conference on
  Robotics and Automation (ICRA)}.\hskip 1em plus 0.5em minus 0.4em\relax IEEE,
  2021.

\bibitem{Lee2015}
R.~Lee, M.~J. Kochenderfer, O.~J. Mengshoel, G.~P. Brat, and M.~P. Owen,
  ``Adaptive stress testing of airborne collision avoidance systems,'' in
  \emph{2015 IEEE/AIAA 34th Digital Avionics Systems Conference (DASC)}.\hskip
  1em plus 0.5em minus 0.4em\relax IEEE, 2015.

\bibitem{Corso2020Survey}
A.~Corso, R.~Moss, M.~Koren, R.~Lee, and M.~Kochenderfer, ``A survey of
  algorithms for black-box safety validation of cyber-physical systems,''
  \emph{Journal of Artificial Intelligence Research}, vol.~72, 2021.

\bibitem{Ding2020}
W.~Ding, B.~Chen, M.~Xu, and D.~Zhao, ``Learning to collide: An adaptive
  safety-critical scenarios generating method,'' in \emph{2020 IEEE/RSJ
  International Conference on Intelligent Robots and Systems (IROS)}.\hskip 1em
  plus 0.5em minus 0.4em\relax IEEE, 2020.

\bibitem{corso2019adaptive}
A.~Corso, P.~Du, K.~Driggs-Campbell, and M.~J. Kochenderfer, ``Adaptive stress
  testing with reward augmentation for autonomous vehicle validatio,'' in
  \emph{2019 IEEE Intelligent Transportation Systems Conference (ITSC)}.\hskip
  1em plus 0.5em minus 0.4em\relax IEEE, 2019, pp. 163--168.

\bibitem{sutton1992reinforcement}
R.~S. Sutton, A.~G. Barto, and R.~J. Williams, ``Reinforcement learning is
  direct adaptive optimal control,'' \emph{IEEE Control Systems Magazine},
  vol.~12, no.~2, pp. 19--22, 1992.

\bibitem{amodei2016concrete}
D.~Amodei, C.~Olah, J.~Steinhardt, P.~Christiano, J.~Schulman, and D.~Man{\'e},
  ``Concrete problems in ai safety,'' \emph{arXiv preprint arXiv:1606.06565},
  2016.

\bibitem{Berlin2021}
J.~Berlin, G.~Hess, A.~Karlsson, W.~Ljungbergh, Z.~Zhang, K.~Åkesson, and
  P.-L. Götvall, ``Trajectory generation for mobile robots in a dynamic
  environment using nonlinear model predictive control,'' in \emph{2021 IEEE
  17th International Conference on Automation Science and Engineering (CASE)},
  2021, pp. 942--947.

\bibitem{schulman2017proximal}
J.~Schulman, F.~Wolski, P.~Dhariwal, A.~Radford, and O.~Klimov, ``Proximal
  policy optimization algorithms,'' \emph{arXiv preprint arXiv:1707.06347},
  2017.

\bibitem{baek2023uncertainty}
W.-J. Baek, T.~P. Huck, J.~Haas, J.~Lewandrowski, T.~Asfour, and T.~Kr{\"o}ger,
  ``Uncertainty-aware risk assessment of robotic systems via importance
  sampling,'' \emph{arXiv preprint arXiv:2308.14068}, 2023.

\end{thebibliography}
\bibliographystyle{IEEEtran}
\end{document}